# Frequency Hopping Synchronization by Reinforcement Learning for Satellite Communication System

Inkyu Kim[1], Sangkeum Lee[2], Haechan Jeong[3], Sarvar Hussain Nengroo[3], and Dongsoo Har[3,*], Senior Member, IEEE

[1]Korea Aerospace Research Institute (KARI), 169-84 Gwahak-ro, Yuseong-gu, Daejeon 34133, Republic of Korea

[2]Department of Computer Engineering, Hanbat National University, 125, Dongseo-daero, Yuseong-gu, Daejeon, Republic of Korea

[3]The CCS Graduate School of Mobility, Korea Advanced Institute of Science and Technology (KAIST), Daejeon 34051, Republic of Korea

**Abstract:** Satellite communication systems (SCSs) used for tactical purposes require robust security and anti-jamming capabilities, making frequency hopping (FH) a powerful option. However, the current FH systems face challenges due to significant interference from other devices and the considerable path loss inherent in satellite communication. This misalignment leads to inefficient synchronization, crucial for maintaining reliable communication. Traditional methods, such as those employing long short-term memory (LSTM) networks, have made improvements, but they still struggle in dynamic conditions of satellite environments. This paper presents a novel method for synchronizing FH signals in tactical SCSs by combining serial search and reinforcement learning to achieve coarse and fine acquisition, respectively. The mathematical analysis and simulation results demonstrate that the proposed method reduces the average number of hops required for synchronization by 58.17% and mean squared error (MSE) of the uplink hop timing estimation by 76.95%, as compared to the conventional serial search method. Comparing with the early late gate synchronization method based on serial search and use of LSTM network, the average number of hops for synchronization is reduced by 12.24% and the MSE by 18.5%.

## I. INTRODUCTION

Satellite communication systems (SCSs) can transmit information over long distances without being limited by geographical boundaries. This technology has become essential in both military and civilian applications, such as command and control, meteorology, remote sensing, and video broadcasting. Generally, a SCS consists of a space-based backbone network, a space-based access network, and a ground backbone network. High-orbit satellites serve as the primary nodes within the space-based backbone network; they possess some computation and storage capabilities and can directly communicate with ground terminals. However, their distance from Earth leads to significant communication delays. The space-based access network includes components like broadband, mobile, and low-orbit satellite networks, which primarily rely on inter-satellite links to facilitate data exchange between the space-based and ground backbone networks. Meanwhile, the ground backbone network, encompassing both mobile and fixed access networks, handles data from space-based satellites and other ground-based sub-networks. As satellite communication systems have evolved, they have become indispensable, offering unparalleled convenience and becoming a crucial part of modern life. SCSs for tactical use prioritize security and anti-jamming capability, making the frequency hopping (FH) spread spectrum technique the preferred choice [1, 2]. The FH is particularly advantageous in tactical scenarios because it rapidly switches the carrier frequency over a wide range of frequencies in a pseudorandom sequence, making it difficult for adversaries to intercept or jam the communication. It can also be used for wireless sensor networks [3] when robust communication becomes an issue. The signal appears as noise unless the specific hopping pattern is known, providing a strong layer of security. In addition, FH helps to mitigate narrowband interference and reduces the effects of multipath fading, which are prevalent in dynamic and hostile environments. These benefits make FH an ideal technique for maintaining secure, reliable communication under challenging conditions. In FH-frequency division multiple access (FH-FDMA) modes for geosynchronous relay satellites, dehop-rehop transponder (DRT) is an economical option at the intermediate frequency stage compared to more expensive baseband processing transponders. However, using different hopping sequences for the uplink and downlink can complicate synchronization between ground equipment and the DRT. Therefore, an efficient synchronization method is needed for FH-FDMA mode using the DRT, specifically for hop synchronization of uplink and downlink.

Wireless communications are extensively used in both civilian and military applications, including 5G, Bluetooth, ultra wide band, satellite communications, and radar systems [4] . However, the inherent broadcast nature of these wireless technologies makes them susceptible to various security threats, particularly malicious jamming attacks. FH spread spectrum (FHSS) is a well-established method for countering such attacks [5]. It is widely employed in both military and consumer communications due to its high energy efficiency. In FHSS systems, the transmitter and receiver synchronize using a shared frequency-hopping pattern to select the carrier frequency for the transmitted signal. This approach makes it difficult for jammers to track legitimate signals, especially if the hopping rate is sufficiently fast.

In recent years, various FH based anti-jamming techniques have been developed to address jamming threats in wireless communications. For instance, adaptive FH (AFH) scheme was introduced in [6, 7] to mitigate mutual interference in multi-user environments by allowing users to dynamically select hopping sets that avoid jammed channels. A study in [8] showed that a differential FH (DFH) system with sequence detection for multi-user high-frequency communications outperformed traditional FH systems. The rise of technologies like software-defined radio and reconfigurable intelligent reflecting surfaces has made it easier to deploy advanced jamming tactics, such as disguised, follower, and reactive jammers [9]. The challenge of anti-jamming communication without pre-shared secrets was first addressed in [10], where an uncoordinated FH (UFH) scheme was introduced. Building on this, author in [11] formulated the UFH-based anti-jamming issue as a non-stochastic multi-armed bandit problem and developed an adaptive UFH algorithm to enhance anti-jamming performance without relying on pre-shared secrets. Various coarse and fine acquisition methods to overcome the jamming issues have been explained in [12, 13]. FH synchronization of DRT-based systems has been examined in [14]. In [12], a ground-triggered FH synchronization method is proposed for relay satellites with the DRT. This method achieves simultaneous FH synchronization for uplink and downlink signals. It uses a non-FH fixed frequency pilot signal transmitted from the ground to synchronize the downlink at the beginning of the synchronization phase. However, this method is vulnerable to malicious uplink jamming and can take longer to achieve final FH synchronization. In addition, compensation schemes for the drift of the satellite can be complicated. The synchronization process for FH communication systems consists of coarse and fine acquisition.

Long short-term memory (LSTM) network and graph convolutional network (GCN) have been employed for the synchronizing tactical communication systems [15-18]. These methods have proven to be more efficient than those described in [12], especially in terms of reducing mean acquisition time (MAT) for synchronization. This enhanced efficiency can be attributed to the ability of the LSTM network to capture long-term dependency and broader context within the signal during fine acquisition. However, the intricate temporal dependencies and bidirectional information flow have led to the exploration of alternative approaches. To overcome the above issues, reinforcement learning (RL) has been introduced as a solution for synchronization due to its ability to adapt in real-time and make sequential decisions based on interactions with a dynamic environment. RL is particularly well-suited for complex and dynamic environments. Unlike traditional methods that rely on fixed rules or prior knowledge and often fail to generalize in constantly changing environments, RL continuously improves its performance by learning from past actions. This allows it to adapt to varying signal conditions, noise levels, and delays, making RL especially valuable for fine acquisition during synchronization, where real-time adjustments are essential for maintaining reliable communication. In addition, the ability of RL algorithms to learn directly from interaction data without needing pre-labeled examples, ensures that the system remains robust even when facing new and unforeseen challenges [19]. In particular, the proximal policy optimization (PPO) algorithm effectively balances exploration and exploitation by optimizing a clipped objective function, preventing large, destabilizing updates to the policy. This stability is crucial in highly dynamic environments like SCSs. The ability of the PPO algorithm to quickly adapt to changing conditions makes it particularly suitable for the fine acquisition phase of synchronization, where real-time performance is critical.

The proposed method is developed to address these significant challenges in synchronizing FH signals in SCSs. Traditional synchronization methods have limitations in terms of efficiency, accuracy, and adaptability to dynamic and noisy environments. This paper introduces a novel method that combines a serial search for coarse acquisition with RL algorithm for fine acquisition. The goal is to significantly reduce both the MAT and mean squared error (MSE) of uplink hop timing estimation, enhancing the overall reliability and security of tactical satellite communication systems. While traditional methodologies, like the LSTM network described rely heavily on supervised training, our approach takes a distinctive path by utilizing the RL technique. The choice of RL is particularly well-suited for scenarios where comprehensive datasets for every possible situation cannot be easily collected. In satellite communications, unpredictable nature of the environment and the necessity for the adaptability in real-time make the adoption of RL particularly appealing. This is primarily because RL excels in situations demanding real-time decisions in environments that are constantly changing. Furthermore, the RL algorithm is applied to problems where it determines a sequence of sequential actions based on a given state in a particular environment. The agent of RL navigates the dynamic energy signal trends, adapting its strategies for synchronization. The acquired knowledge allows the agent to execute fine acquisition even in the presence of significant noise. The main contributions of this study are:

1) This paper presents a novel synchronization technique that integrates a serial search for coarse acquisition with PPO algorithm for fine acquisition, aimed at reducing the MAT and MSE in uplink hop timing estimation.

2) The proposed method improves accuracy and demonstrates better adaptability in highly dynamic environments. The combination of RL and PPO ensures that the synchronization process remains efficient, even under challenging conditions such as fluctuating signal-to-noise ratios (SNR).

3) In addition, scalability of proposed method makes it applicable to a wide range of satellite communication scenarios, ensuring reliable synchronization without the need for extensive retraining or prior knowledge of the environment.

The remainder of this study is organized as follows. Section II explains the operation of the DRT within the tactical SCS. The RL-based scheme is outlined in Section III. Section IV presents the simulation results, and Section V provides the concluding remarks.

## II. FH-FDMA MODEL FOR SYNCHRONIZATION SYSTEM

Fig. 1 shows the functional diagram of the FH synchronization process within the tactical SCS. The key components in this communication loop include a DRT in the relay satellite, a synchronization modem (SM) at the ground station, comprising transmitter ($T_x$) and receiver ($R_x$), and a FH synchronizer controlled by a timing controller [12, 18]. The FH signal $r^u(t)$ received from the SM is dehopped at the DRT and converted to the intermediate frequency (IF) band for further processing, including group dehopping or rehopping. Three key elements are required to generate the random frequencies for dehopping at the DRT: the transmission security key $k^{UP}$ from the key supplier, the clock $t^G$ generated by the clock generator, and the time of day *(TOD)* provided by TOD counter [2], [16]. Random frequency tables are created using $k^{UP}$, while indices needed to access these tables are generated by combining *TOD* and $t^S$. With the shared $k^{UP}$ between the satellite and the ground station, the DRT uses *TOD* and $t^S$ to dehop the signal transmitted by Tx, allowing it to recover the signals corresponding to each frequency slot. These recovered signals then pass through a bandpass filter, grouping dehopping and rehopping. Finally, the DRT transmits the processed signals back to the ground station after reallocating them to new set of random frequencies through rehopping process, using the downlink key $k^{DN}$.

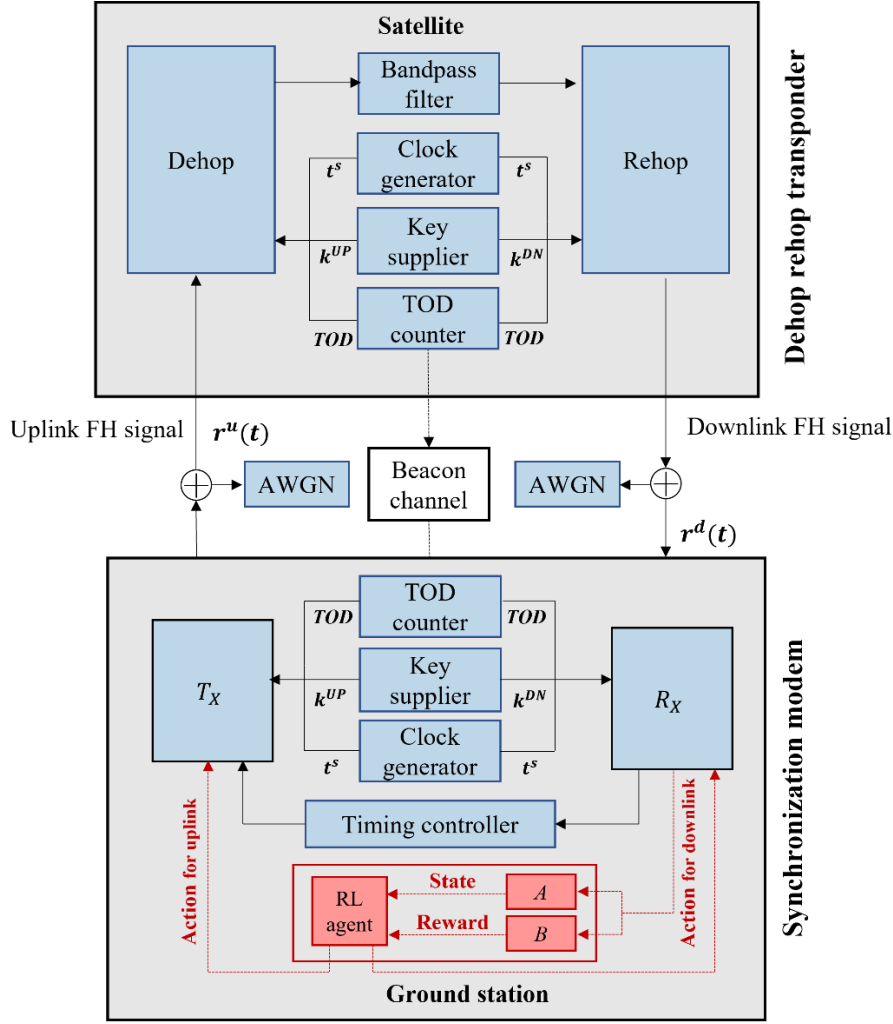

Figure 1. Functional block diagram of SCS. The SCS achieves synchronization by using a timing controller within the ground station, as indicated by the black solid lines. The red dot lines represent the synchronization process by the proposed RL algorithm. The two functional blocks labeled as A and B generate the time-varying squared magnitude of the dehopped downlink signal representing current state and the average of squared magnitude of the dehopped downlink signal representing reward.

The signals received by the receiver $R_x$ are restored using the pre-shared $k^{DN}$ and $TOD$ obtained via the beacon channel. FH is achieved by randomly selecting frequencies at the transmitter, decrypting them at the satellite, reassigning them to new frequencies, and finally restoring the original signal at the ground station. However, as signals are transmitted from the $T_x$ and restored at the $R_x$ through both uplink and downlink, they experience quality degradation due to delays in each stage of the process. To address this, synchronization is managed by the timing controller. The frequencies used for uplink communication generated with $k^{UP}$ are distinct from those used for downlink communication generated with $k^{DN}$. This distinction allows the receiver to measure uplink and downlink delays separately. Based on these measurements, the timing controller adjusts synchronization for each, enabling fast and efficient system synchronization. The mathematical expressions defining the variables used for explaining the synchronization process are adopted from [12]. For uplink communication with the satellite, the *jth* frequency hop, including additive white Gaussian noise (AWGN) $n^u(t)$ and time-varying propagation delay $\tau_u$, received at the DRT can be expressed as

$$r^u(t) = \sum_j A\omega_j(t) \cos(2\pi f_j^u t)(t - \tau_u) + n^u(t) \qquad (1)$$

where $A$ is the amplitude, $\omega_j(t)$ is a shaping pulse in the *j*th hop in the hop duration, and $f_j^u$ is the frequency index for the *j*th hop for uplink signal.

The dehopped signal is obtained by processing the received signal through the dehopper using the *k*th frequency. The corresponding bandpass-filtered signal, represented as $b_k^u(t)$, can be given as

$$b_k^u(t) = BP[r^u(t) w_k^u(t) w_{LO}^u(t)] \qquad (2)$$

where $w_k^u(t)$ denotes pulse shaping for the *kth* frequency and $w_{LO}^u(t)$ indicates the selected frequencies by the indices from the pseudonoise sequence generator. The $w_k^u(t)$ and $w_{LO}^u(t)$ are provided to the frequency synthesizer and converted to the local oscillator output. In the downlink communication, the received signal $r^d(t)$ at the R$_x$ of the SM can be given by

$$r^d(t) = \sum_l b_k^u(t)\, \omega_l(t) \cos(2\pi f_j^u t)(t - \tau_d) + n^d(t) \qquad (3)$$

where $\omega_l(t)$, $f_j^d$, $\tau_d$, and $n^d(t)$ are the shaping pulse in the *l*th hop with the hop duration, the frequency index for the *l*th hop for downlink signal, time-varying propagation delay, and AWGN noise, in the downlink communication, respectively.

When considering the unknown process delay $\tau_h$ resulting from the operation of the DRT, the total delay can be expressed as $\tau_u + \tau_d + \tau_h$. The process delay $\tau_h$ can be assumed constant and negligible as compared to $\tau_u$ and $\tau_d$. The $\tau_u$ of uplink and $\tau_d + \tau_h$ of downlink are compensated by the timing controller for synchronization. It is shown in [12] that the number of hop durations required for synchronization viewed by T$_x$ is $(N_d + N_u) + \lambda_u$, where $(N_d + N_u)$ represents number of hop durations corresponding to $\tau_d + \tau_u$ and $\lambda_u$ indicates integer random variable to be adjusted by the T$_x$ for coarse acquisition. When viewed by R$_x$, it is $(N_d + N_u) + \lambda_d$, where $\lambda_d$ represents integer random variable to be adjusted by the R$_x$.

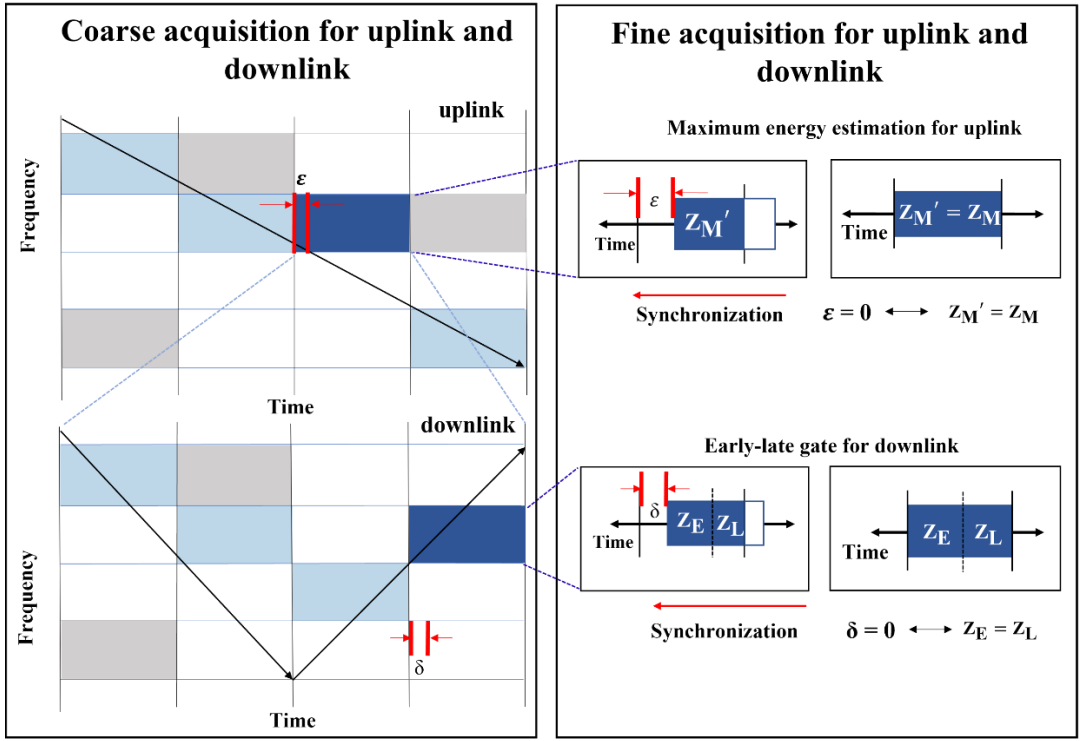

Figure 2. Synchronization scheme for uplink and downlink signals. In the time-frequency coordinate system, the gray rectangles represent the actual FH signals. The light blue rectangles are the cells targeted for the serial search process, and the dark blue rectangles represent the cells that match with the actual FH signal. Coarse acquisition is achieved by the serial search represented by thick black lines, while fine acquisition is accomplished by using the maximum energy estimation for uplink and early-late gate (ELG) method for downlink. $Z_{M'}$, $Z_M$, $Z_E$, $Z_L$ represent energy over symbol or half-symbol time duration. RL algorithm is employed in this work for fine acquisition in both uplink and downlink instead of the maximum energy estimation and the ELG.

Fig. 2 illustrates the synchronization process for both uplink and downlink communication. To reduce payload costs, the satellite is not equipped with synchronization hardware. Therefore, it is crucial for the transmitter to be precisely synchronized when sending the randomly generated FH signals during the uplink. Similarly, during the downlink process, the receiver must synchronize accurately to receive signals, as delays can occur when the satellite transmits signals. As a result, achieving effective synchronization in both the uplink and downlink processes is crucial for the smooth communication between the ground station and the satellite. During the coarse acquisition, the $T_x$ and $R_x$ use a serial search algorithm. This algorithm begins by searching through the first hopping frequency and continues sequentially through each frequency until a match with the actual hopping frequency is found. As the serial search progresses through all the hopping sequences, the receiver detects the start of the hopping symbol. It measures the energy of the received signal over the symbol duration, and stores the highest energy value as the maximum energy value ($Z_M$). The $Z_M$ represents the accumulated energy over a symbol duration when the serial search detects the actual beginning of FH signal, i.e., the time offset $\varepsilon$ in Fig. 2 is 0. Once a hopping frequency is found by the serial search that produces $Z_M$, the $R_x$ uses that hopping symbol for fine acquisition.

To ensure precise communication, fine acquisition is essential for minimizing synchronization errors. In the uplink process, fine acquisition is achieved using maximum energy estimation. The Rx calculates the energy received over a symbol duration, denoted by $Z_M'$, and Tx adjusts ε until $Z_M'$ matches the maximum value $Z_M$. When these two values are equal, uplink synchronization is achieved. For the downlink, the fine acquisition process utilizes the ELG method. This method compares the energy value obtained during the first half of a hop duration, denoted by $Z_E$, with the energy value obtained during the second half of a hop duration, denoted by $Z_L$. The Rx adjusts the time offset δ until $Z_E$ and $Z_L$ are equal. When these values match, downlink synchronization is achieved.

### III. PROPOSED FH SYNCHRONISATION METHOD USING RL AND GCN-BI-LSTM NETWORK

RL is a growing field aimed at improving cognition in various systems. Unlike supervised learning, RL doesn't require labeled data to learn. A core principle of RL is "learning from interactions." Figure 1 illustrates the framework of episodic RL. In this framework, an agent interacts with its environment over a series of time steps. At each step $t$, the agent observes the environment's current state and takes an action based on that observation, which then causes the environment to transition to a new state at the next time step. At the terminal state, marking the end of an episode, the environment provides the agent with a reward $R$. The proposed method employs serial search for coarse acquisition and RL for fine acquisition. Markov Decision Process (MDP), a mathematical framework is used to model decision-making problems where the outcome of an action is uncertain. In the SCS, the MDP can be used to optimize the system performance by identifying the optimal sequence under uncertainty conditions. The MDP models the decision-making process by defining state, action, and reward, used as the foundation for the RL framework. The environment depicted in Figure 3 represents the synchronization process within the SCS.

#### A. PPO for Actor-Critic with GCN-Bi-LSTM Network

In RL, the PPO algorithm is fundamental and widely utilized method for achieving optimal policy learning. This study leverages the PPO algorithm as the core foundation of RL framework. The PPO algorithm offers a substantial advantage over the slow convergence of the deep Q-learning algorithm in training, leading to faster convergence and higher performance rate than other RL algorithms. For RL, the actor-critic structure is adopted. The actor's primary function is to estimate the policy, which determines the agent's actions in a given state, and the critic focuses on estimating the value function, which predicts the expected future reward for a particular state or state-action pair. In addition, the actor's policy is refined based on the feedback from the critic.

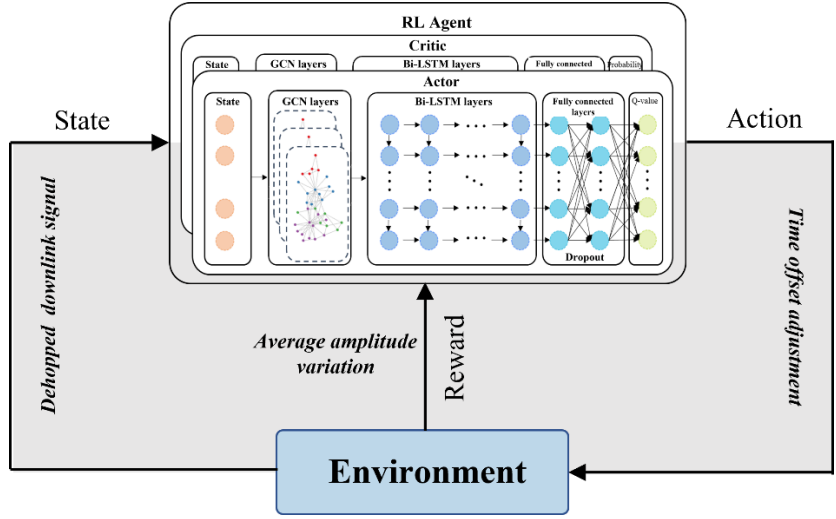

Figure 3. RL framework for fine acquisition, e.g., interaction between environment and RL agent structured into actor and critic components. Actor and critic components are respectively implemented by using the GCN-Bi-LSTM network.

In this study, the actor plays a pivotal role by utilizing the state and Q-value as its input parameters, and subsequently produces an output in the form of an action. This action encompasses various adjustments, such as modifying the time offset ε/δ for the uplink/downlink signal. On the other hand, the critic takes the state and reward as its input parameters, and generates the Q-value output serving as a critical metric for assessing the performance. In PPO, the policy followed by the RL actor is defined by the combination of networks. Specifically, the actor network outputs a probability distribution over possible actions, representing the policy. Unlike purely exploitation-based approaches where the action with the highest Q-value would always be chosen, the actor samples actions from this distribution. This stochastic policy allows a balance between exploration (trying new actions) and exploitation (choosing actions that are known to be effective). This balance is critical in ensuring that the agent can avoid local optima and adapt to the dynamic nature of the satellite communication environment. The critic network estimates the value function, which guides the actor's policy updates by providing feedback on the expected rewards of actions taken. While the actor's decisions are influenced by the critic's value estimates, the final action is probabilistically determined by the actor's policy, ensuring both robustness and adaptability. The DQN technique introduced in [20], integrates the traditional Q-Learning with neural networks [21, 22]. This combination allows DQN to efficiently approximate functions, enabling faster convergence of action-value functions with fewer training iterations [23]. DQN is an extension of classical Q-Learning, designed to approximate the optimal action-value function, represented as

$$Q(s,a) = \gamma \, max\big(Q(s',a')\big) + r(s,a) \tag{4}$$

where $Q(s,a)$ is represents the Q-value of taking action $a'$ in next state $s'$. The discount factor $\gamma \in [0,1)$ determines the importance of rewards received over time, the reward function $r(s,a)$ gives the reward for performing action $a$ in current state $s$. The $max(Q(s',a'))$ represents the maximum possible Q value of the next state $s'$ by selecting the optimal action $a'$. The DQN updates the Q-values at each time step using the loss function

$$L(\theta) = \left[\left(r(s,a) + \gamma\, max\big(Q(s',a';\theta)\big)\right) - Q(s,a;\theta)\right]^2 \qquad (5)$$

where $L(\theta)$ is the loss function taking policy parameter vector $\theta$. The term $\left(r(s,a) + \gamma\, max\big(Q(s',a';\theta)\big)\right)$ represents the target Q value, while $Q(s,a;\theta)$ is the predicted Q value. The network of RL is trained by minimizing $L(\theta)$.

PPO is a policy-gradient approach that does not rely on a model and follows an on-policy and actor-critic architecture. Its goal is to preserve the trustworthy performance of TRPO algorithm that ensure monotonic improvements by considering the KL divergence of policy updates, while using only first-order optimization. To improve sample efficiency, PPO employs importance sampling to estimate the expected value of samples collected from a previous policy under the updated policy. This allows each sample to be used for multiple gradient iterations. However, as the updated policy evolves, it will diverge from the previous policy, leading to an increase in estimation variance. To address this issue, the policy is regularly updated to match with the updated policy. For this technique to be considered valid, the state transition function must be similar for both policies. To ensure this, PPO constrains the probability ratio in (3a) within the range of $[1 - \epsilon_{clip}, 1 + \epsilon_{clip}]$ through a process called clipping. This approach assures the similarity of state transition functions and provides a first-order method for optimizing the trust region. The PPO is similar to TRPO, but it uses an alternative surrogate objective function designed to be more straightforward to implement [24]. The clipped surrogate objective $L^{CLIP}(\theta)$ taking policy parameter vector $\theta$ is given as

$$L^{CLIP}(\theta) = \widehat{\mathbb{E}}_t\left[min\big(r_t(\theta)\hat{A}_t, clip(r_t(\theta), 1 - \epsilon_{clip}, 1 + \epsilon_{clip})\hat{A}_t\big)\right]$$

$$= \widehat{\mathbb{E}}_t\left[min\left(\frac{\pi_\theta(a_t|s_t)}{\pi_{\theta_{old}}(a_t|s_t)}\hat{A}_t, clip\left(\frac{\pi_\theta(a_t|s_t)}{\pi_{\theta_{old}}(a_t|s_t)}, 1 - \epsilon_{clip}, 1 + \epsilon_{clip}\right)\hat{A}_t\right)\right] \qquad (6)$$

where $\epsilon_{clip}$ is a hyperparameter which is used to prevent moving $r_t$ outside of the interval $[1 - \epsilon_{clip}, 1 + \epsilon_{clip}]$.

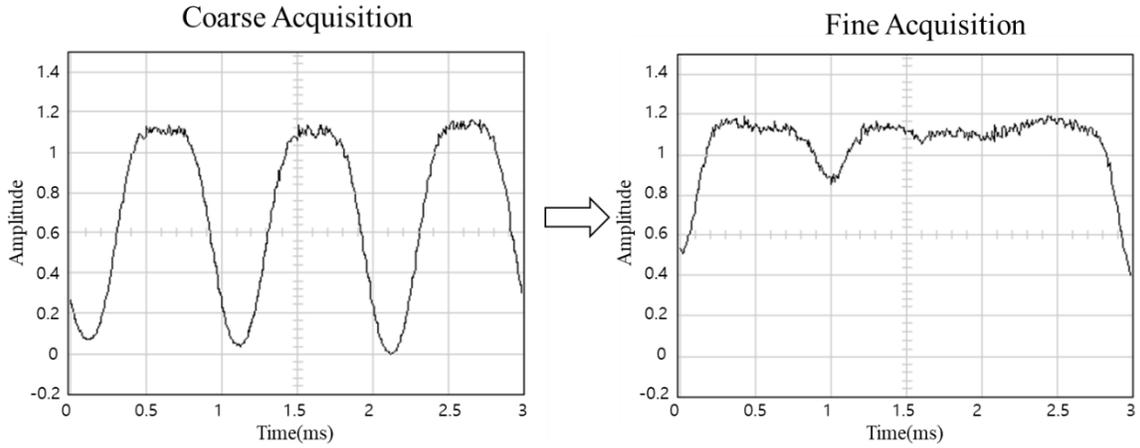

(a)

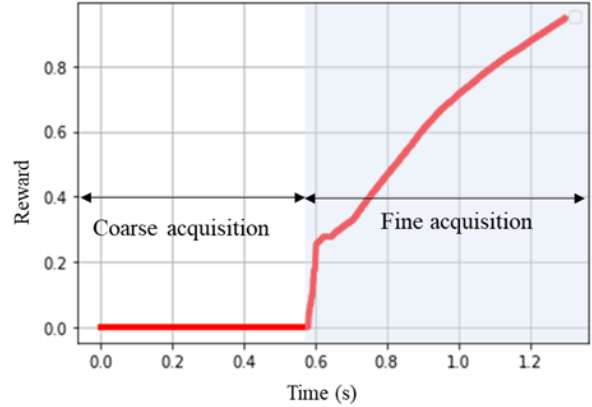

(b)

Figure 4. Time-varying squared magnitude of dehopped downlink signal and reward: (a) time-varying squared magnitude of dehopped downlink signal obtained from coarse acquisition (left plot) and fine acquisition (right plot); (b) reward evaluated by averaging time-varying squared magnitude of 50 dehopped signal samples obtained after taking time-offset adjustment action.

To enhance predictive performance in complex operational scenarios involving engineered systems, the Bi-LSTM network is utilized. The Bi-LSTM network has a dual-layer structure, featuring two main hidden layers. One layer processes information forward, while the other processes it in reverse. Unlike a standard LSTM, Bi-LSTM can capture both past and future data by employing two LSTM cells that work in opposite directions [25]. Each GCN-Bi-LSTM network consists of 2 GCN layers, 5 Bi-LSTM layers, and 3 fully-connected layers. In this study, a GCN-Bi-LSTM network is employed within an actor-critic architecture to enhance the performance of PPO-based training for fine acquisition, as depicted in Fig. 3. The GCN-Bi-LSTM model is designed to process time-varying data, transforming temporal relationships into spatial relationships for improved analysis. The input data to the GCN layer is the time-varying squared magnitude of the dehopped downlink signal. This magnitude represents the state of fine acquisition. To map this state information into graph data of GCN processing, each point in the time sequence of the dehopped signal becomes a node in the graph. Edges between nodes are then established on their temporal proximity. This transformation enables the network to convert temporal relationships from the original sequence into spatial relationships within the graph. Each node in the GCN layer updates its information by incorporating information from adjacent nodes, allowing the GCN layer to extract patterns representing the relationships among sampled time intervals within the dehopped downlink signal. The output is passed to the Bi-LSTM network. The Bi-LSTM network effectively captures past and future contextual information, enabling a more comprehensive understanding of sequential data. The actor-critic architecture uses the GCN-Bi-LSTM network for decision-making, with the actor estimating the policy and the critic evaluating the value function. In addition, the Bi-LSTM network can capture sequential dependencies among signals over different time intervals, emphasizing the long-term dependency of the squared magnitude of the dehopped downlink signal. Therefore, the benefit stems from the synergy between GCN and Bi-LSTM networks, allowing the simultaneous incorporation of local information within time series data and global patterns across the dataset.

| | |
|---|---|
| **Algorithm 1** PPO algorithm for FH | |
| 1: | **Initialize environment:** Set PPO hyperparameters, γ; clip ratio, search region, $R$; dwell time, and sample size $N = 50$; estimated advantage, $\hat{A}_t$ |
| 2: | Initialize time shift interval $I$, hop duration $\tau_h$, uplink offset $\varepsilon$, downlink offset $\delta$ |
| 3: | Initialize policy parameters, $\theta$; number of time steps per episode, $T$; state at time t, $s_t$; action taken at time t, $a_t$; reward received after taking action $a_t$ in state $s_t$, $R_t$; policy function, $\pi$; set of trajectories at iteration $k$, $D_k$; |
| 4: | Coarse acquisition using serial search |
| 6 | Fine acquisition using RL |
| 7 | Initialize PPO actor-critic networks |
| 8 | Initialize state (7.1) with dehopped signal after coarse acquisition |
| 9 | Select action from (7.2) |
| 10 | Apply action and update environment |
| 11 | Update state based on selected action and compute new time offsets (ε, δ) |
| 12: | Calculate reward from (7.3) |
| 13: | Store state, action, and reward |
| | Compute MAT from (8) |
| 14: | Compute MSE for uplink hop timing estimation from (9) |
| 15: | **end for** |

*B. State, Action, and Reward*

The RL framework shown in Fig. 3 consists of two components: the learning agent (RL agent) and the synchronization process (environment). During each time interval, the agent observes the current state of the environment, chooses an action, and executes it, thereby altering the state of the environment and receiving a corresponding reward. A set of 50 samples that together make up the state gives the condition of synchronization. As shown in Fig. 4(a), each of these samples corresponds to an empirical assessment of the squared magnitude of the dehopped downlink signal. Actions derived from this state are altering time offsets $\varepsilon$ and $\delta$. Every action results in a subsequent state, representing the system's new synchronization status. Reward, which plays a significant role in the agent's behaviors, serves as a quantitative assessment of action. In the proposed method, the reward is determined as the mean squared magnitude of 50 dehopped downlink signal samples in the subsequent state in Fig. 4(b). State, action, and reward can be expressed as follows

$$state = \begin{bmatrix} \{r^d(t_1) \exp(-j2\pi f t_1)\}^2, \\ \{r^d(t_2) \exp(-j2\pi f t_2)\}^2, \\ \dots \\ \{r^d(t_N) \exp(-j2\pi f t_N)\}^2 \end{bmatrix} \quad (7.1)$$

$$action = \begin{cases} T_{offset} \leftarrow T_{offset} - I, & T_{offset} > 0 \\ T_{offset} \leftarrow T_{offset} + I, & T_{offset} < 0 \\ T_{offset} = T_{offset}, & T_{offset} = 0 \end{cases} \quad (7.2)$$

$$reward = \frac{1}{N}\sum_{i=1}^{N}\{r^d(t_i)\exp(-j2\pi f t_i)\}^2 \quad (7.3)$$

where $T_{offset}$ represents the uplink time offset ε and downlink time offset δ, $I$ denotes the time shift interval and $N$ represents the number of samples set to 50.

When the reward closely approaches its maximum value, as depicted by the flat peak near 1.2 in Fig. 4(a), it serves as a clear indicator of achieving the synchronization. It is important to note the interdependency between uplink and downlink signals. Alterations in uplink timing (ε) are inherently linked to the variations observed in the downlink signal.

## IV. SIMULATION RESULTS

To evaluate the effectiveness of the proposed coarse acquisition scheme, we use the MAT. Although the MAT varies with the search strategy applied, we define it in accordance with equation (8), based on the serial search method [14]:

$$\bar{T}_a = (R-1)\left(\frac{2-P_D}{2P_D}\right)\bar{T}_{ic} + \left(\frac{1-P_D}{P_D}\right)(\bar{T}_c + T_r) + \bar{T}_{ac} \quad (8)$$

where $R$ is the uncertainty region, $P_D$ is the probability of correctly detecting the target cell during a scanning period within the uncertainty region. $\bar{T}_{ic}$ is the expected time to dismiss an incorrect cell, while $\bar{T}_c$ represents the expected time to dismiss the correct cell. $T_r$ is the rewinding time required to return to the first cell in the uncertainty region, and $\bar{T}_{ac}$ is the expected time for accepting the correct cell.

In addition, MSE of uplink hop timing estimation is also used to evaluate the performance of the fine acquisition in the uplink signal and is defined as

$$MSE\left(\frac{\hat{t}_h}{T_h}\right) = E\left[\left(\frac{\hat{t}_h - \tau_h}{T_h}\right)^2\right] = \frac{1}{N}\sum_{m=1}^{N-1}\left(\frac{\hat{t}_{h,m} - \tau_{h,m}}{T_h}\right)^2 \quad (9)$$

where $\hat{t}_h, \tau_h, T_h$ are estimated uplink hop timing, true uplink hop timing, and the hop duration, respectively. The hop timing error $\hat{t}_h - \tau_h$ indicates the timing difference between the uplink FH signal and reference FH signal for dehopping.

The simulation parameters, excluding the PPO hyperparameters listed in Table II, are consistent with those used in [12] [18], and are listed in TABLE I. The satellite operates in a geosynchronous orbit at an altitude of approximately 35,786 km above the Earth's equator. This orbit allows the satellite to remain stationary relative to the Earth's surface, enabling continuous communication with ground stations. Given the significant distance between the satellite and the ground stations, the model accounts for a uniformly distributed propagation delay for both uplink and downlink communications. The propagation delay for the uplink is modeled within the range of 59.80 to 60.65 milliseconds, while the downlink delay ranges from 59.35 to 60.25 milliseconds. These delays are critical for accurately reflecting the timing challenges involved in synchronization. The carrier frequency used in the simulations is set within the X-band (7.25 GHz to 8.4 GHz), which is commonly used for military and tactical satellite communications due to its resilience against interference and its ability to penetrate through various atmospheric conditions. The satellite model also includes FH across a wide range of frequencies within the X-band, with the FH pattern generated pseudorandomly. The hop duration is set to 1 millisecond, allowing the system to switch frequencies quickly and improve security and anti-jamming capabilities. The satellite is equipped with directional antennas providing a gain of approximately 30 dBi. This high gain is necessary for directing the signal toward specific ground stations, minimizing signal loss, and enhancing communication reliability. The SNR in the satellite communication link is modeled within the range of 0 to 20 dB covering conditions from poor to excellent signal quality, reflecting the challenges of real-world tactical communication. The satellite communication model uses Offset Quadrature Phase Shift Keying (OQPSK) as the modulation scheme. The OQPSK is chosen for its ability to preserve signal integrity over long distances and remain robust against noise and interference. Even though the satellite remains in a geosynchronous orbit, slight drift caused by gravitational forces is modeled. This drift is corrected by a timing controller that adjusts the synchronization process to accommodate the changes. Dwell time for serial search is set to 20ms and one-hop shift interval for serial search is set to 100ms for uplink and 20ms for downlink. For fine acquisition, the dwell time is 5 milliseconds for the uplink and 1 millisecond for the downlink. The time shift interval for evaluating the ELG and maximum energy estimation is commonly set to 20 microseconds. With a hop duration of 1 millisecond, ZE is calculated over the first 0.5 milliseconds, while ZL is measured over the second 0.5 milliseconds. The hyperparameters for the LSTM network are adopted from [18].

The references used for experimental comparison in this study were selected based on their relevance to FH synchronization in SCSs. Specifically, [18] was chosen because it represents one of the most advanced approaches to synchronization using a LSTM network. The LSTM-based method has been shown to significantly improve the MAT and MSE in uplink hop timing estimation, making it a strong benchmark for evaluating new methods. In addition, [12] and other traditional methods were selected providing a solid baseline for assessing the improvements brought by the proposed RL approach. By comparing the proposed RL method with both advanced and traditional methods, the study aims to demonstrate the effectiveness of the RL approach across a range of established techniques. The input data for RL training consists of the time-varying squared magnitude of the dehopped downlink signal, as shown in Fig. 4(a). Each data point corresponds to a 10μs time interval. The mini-batch size is set to 200, with each mini-batch sampled from the squared magnitude of the dehopped downlink signal collected over a 200-second period. For testing, an additional 300 seconds of time-varying squared magnitude data from the dehopped downlink signal is used.

TABLE I Simulation parameters

| Parameter | Value |
| --- | --- |
| Sample rate | 100kHz |
| Hop duration | 1ms |
| Uniformly distributed propagation delay in uplink | 59.80~60.65ms |
| Uniformly distributed propagation delay in downlink | 59.35~60.25ms |
| SNR | 0~20dB |
| Modulation | OQPSK |

TABLE II Hyperparameters for PPO

| **Hyper parameter** | **Value** |
|---|---|
| Discount factor ($\gamma$) | 0.99 |
| Update interval | 128 |
| Actor learning rate | 0.0005 |
| Critic learning rate | 0.001 |
| Clip ratio | 0.1 |
| Smoothing parameter ($\lambda$) | 0.95 |
| Epochs | 3 |
| Optimizer | Adam |
| Exploration rate ($\epsilon$) | 0.1 |
| Decay rate of exploration | 0.995 |
| Minimum value of exploration rate | 0.01 |

Fig. 4(b) shows the variation of reward. For the time period of coarse acquisition until 0.6s, the reward remains at 0. However, during the fine acquisition after 0.6s, the RL agent learns over time, leading to increased reward value. During training, 0-20dB range of SNR is considered to ensure robust synchronization performance.

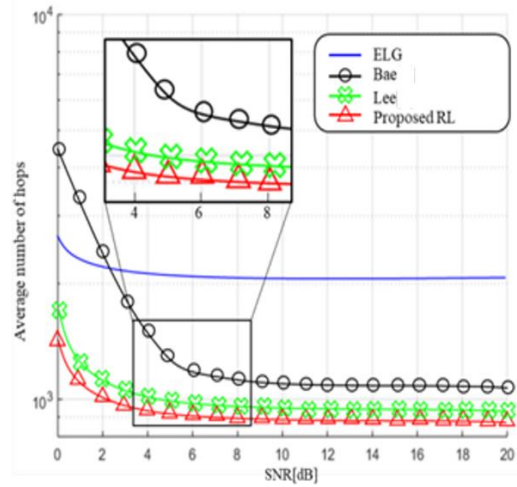

(a)

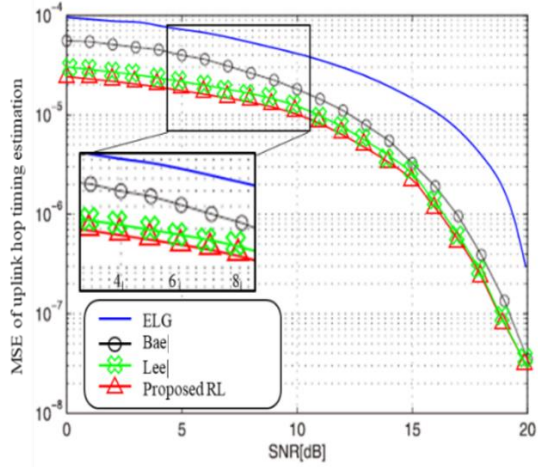

(b)

Figure 5. Comparison of the MAT and MSE of uplink hop timing estimation using four methods with serial search commonly adopted for coarse acquisition: (a) Average number of hops; (b) MSE of uplink hop timing estimation.

Fig. 5(a) shows the average hop count obtained by "ELG", "Bae [12]", "Lee [18]", and the RL method after training. The "ELG" represents the satellite-triggered synchronization. The "ELG" achieves fine acquisition in uplink and downlink commonly by the ELG method. As seen in Fig. 5(a), the RL method achieves the best performance, characterized by the lowest average hop count over the entire range of SNR. In comparison, the ELG method appears to yield the least favorable results. To determine the successful synchronization achieved by the PPO algorithm, the behavior of $Z_M$ and the alignment between $Z_E$ and $Z_L$ are observed. Synchronization by both the PPO and LSTM network is considered complete when $Z_M$ reaches its maximum value, and $Z_E$ equals $Z_L$. This corresponds to the condition of fine acquisition for both links. The MAT is calculated by aggregating the number of hops required by both the serial search and individual fine acquisition techniques. In comparison with the "ELG" method, the MAT of "Bae", "Lee", and the RL method exhibits an average reduction of 29.60%, 52.34%, and 58.17%, respectively, over the 0-20dB SNR range. Moreover, when compared with "Lee", the average number of hops of the proposed RL is decreased by 12.24%.

A lower MSE value indicates a more accurate estimation of the uplink hop timing. Fig. 5(b) shows the MSE values obtained from different synchronization methods. The MSE of "Bae", "Lee", and the RL method evaluated over the 0-20dB range of SNR is decreased by 49.45%, 71.71%, and 76.95%, respectively, as compared to "ELG". This improvement can be attributed to the ability of ground-triggered methods to synchronize the uplink and downlink separately, enabling faster timing adjustments. When compared with "Lee", the MSE of uplink hop timing estimation of the RL method is decreased by 18.50%.

## V. CONCLUSION

This paper introduces a novel synchronization method that combines serial search for coarse acquisition with PPO algorithm for fine acquisition to reduce MAT and MSE of uplink hop timing estimation. The agent of the PPO algorithm can learn the temporal trends of the dehopped downlink signal and achieves faster synchronization. Compared to the results presented in [8], which used an LSTM network, the proposed method reduces MAT by 12.24% and the MSE of uplink hop timing estimation by 18.50% over the 0-20 dB range of SNR. When compared with other methods, the MAT and MSE of uplink hop timing estimation is reduced. However, this paper does not include a direct performance comparison related to advancements in network architecture. Future work focuses on evaluating how different network architectures, such as variations in neural network layers or advanced models like transformer networks, impact the overall performance of the proposed synchronization method. Such a comparison would provide a more comprehensive understanding of the method's effectiveness and scalability in various scenarios.